%% file: main.tex
\documentclass{IEEEtran}

\usepackage{cite}  %
\usepackage{graphics}
\usepackage{epsfig}
\usepackage{times}
\usepackage{amsmath}
\usepackage{amssymb}
\usepackage{subcaption}
\usepackage{multirow}
\usepackage{siunitx}
\usepackage[ruled,vlined]{algorithm2e}
\usepackage{flushend} 
\usepackage{wrapfig}
\usepackage{pdfpages}
\usepackage{hyperref}

\input{maths_commands}

\title{Task and Joint Space Dual-Arm Compliant Control}
\date{}

\author{
\begin{tabular}{ccc}
Alexander L. Mitchell$^{*1,2}$,
Tobit  Flatscher$^{*2}$,
Ingmar Posner$^{1,2}$
\end{tabular}
\vspace{0.1cm} \\
$^{1}$Applied AI Lab,
$^{2}$Oxford Robotics Institute%
\vspace{-0.75cm}
}

\begin{document}
\maketitle

\def\thefootnote{*}\footnotetext{Equal Contribution; Email: \tt{\{mitch,tobit\}@robots.ox.ac.uk}}\def\thefootnote{\arabic{footnote}}

\input{sections/0_abstract}
\input{sections/1_introduction}

\input{sections/2_methods}

\input{sections/3_Results}

\input{sections/4_conclusions}
\input{sections/5_acknowledgements}

\bibliographystyle{IEEEtran}
\bibliography{references}

\end{document}

%% file: maths_commands.tex
\usepackage{amsmath,amsfonts,bm}

\def\eqref#1{equation~\ref{#1}}

\def\1{\bm{1}}

\DeclareMathAlphabet{\mathsfit}{\encodingdefault}{\sfdefault}{m}{sl}
\SetMathAlphabet{\mathsfit}{bold}{\encodingdefault}{\sfdefault}{bx}{n}

%% file: sections/0_abstract.tex
\begin{abstract}
    Robots that interact with humans or perform delicate manipulation tasks must exhibit compliance. However, most commercial manipulators are rigid and suffer from significant friction, limiting end-effector tracking accuracy in torque-controlled modes. To address this, we present a real-time, open-source impedance controller that smoothly interpolates between joint-space and task-space compliance. This hybrid approach ensures safe interaction and precise task execution, such as sub-centimetre pin insertions.
    We deploy our controller on Frank, a dual-arm platform with two Kinova Gen3 arms, and compensate for modelled friction dynamics using a model-free observer. The system is real-time capable and integrates with standard ROS tools like MoveIt!.
    It also supports high-frequency trajectory streaming, enabling closed-loop execution of trajectories generated by learning-based methods, optimal control, or teleoperation. Our results demonstrate robust tracking and compliant behaviour even under high-friction conditions. The complete system is available open-source at \href{https://github.com/applied-ai-lab/compliant_controllers}{link}.
\end{abstract}

%% file: sections/1_introduction.tex
\section{Introduction}

Compliant robots are crucial for safe operation in the vicinity of people. However, the majority of robot manipulators are rigid and mechanically noncompliant. Therefore, we implement an impedance controller meaning that the manipulator responds like a virtual mass damper system~\cite{siciliano2007Springer}. Impedance control has a long history and is well understood being used for contact-rich manipulation~\cite{Hogan1984Impedance,DeRisiVariable2024,Lu2024Fuzzy,Rysbek2024Proactive,collins2023ramp} and legged locomotion~\cite{towr,bellicoso2018dynamic,boxFDDP} as well as being shipped with some manipulators such as the Frank-Emika Panda~\cite{Haddadin2022Franka}.

We introduce an open-source, impedance controller which blends joint-space and task-space control, enabling real-time transitions between these domains via dynamic reconfiguration. Our controller is designed for safe human interaction, handles complex assembly tasks like pin insertion, and compensates for friction through a model-free observer~\cite{compliant_formulation}—all within a unified and stable control framework. Unlike prior work, our system supports continuous interpolation between compliance modes without restarting or reinitializing the controller. This makes it particularly well-suited for advanced manipulation scenarios that require both precise trajectory tracking and adaptable compliance, such as learning-from-demonstration, optimal control, and teleoperation.

To implement the default controller we combine the joint- and task-space controllers from~\cite{compliant_formulation} into a single controller. This results in complete control of all seven degrees of freedom and low task-space tracking error.
The task- and joint-space tracking error is reduced further using a model-free friction observer, the formulation of which is also presented in \cite{compliant_formulation}.
Additionally, the controller gains can be adjusted in real time using a dynamic reconfiguration node. This allows smooth transitions between task-only, joint-only, and combined control modes, enabling adaptable behaviour depending on the application.
For example, precise and localised motion such as pin insertion or removal can be executed using the controller biased towards task space, whilst large sweeping motions of the end-effectors during reaching tasks are executed using the joint and task-space controller in combination. The combined task and joint space controller remains stable under the passivity criterion~\cite{Tiseo2023Safe}.

\begin{figure}[t]
    \centering
    \includegraphics[width=\linewidth]{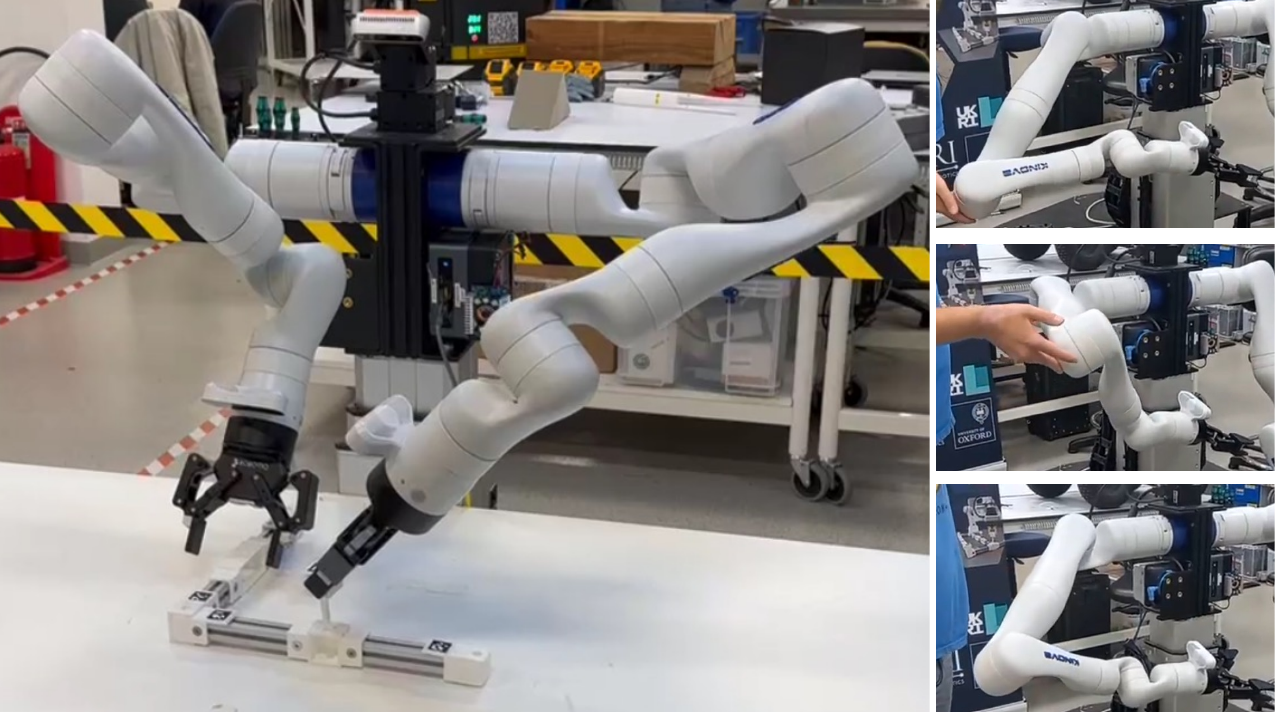}
    \caption{Task- and joint-space compliant control as deployed on our real dual-arm platform christened Frank. The controller is based on~\cite{compliant_formulation}. We combine the joint- and task-space formulations to create a tracking controller precise enough to perform pin insertions, whilst remaining compliant and safe for human-robot interaction. Please find and use our implementation at \url{https://github.com/applied-ai-lab/compliant_controllers}.}
    \label{fig:first-figure}
    \vspace{-0.5cm}
\end{figure}

Our contributions is an open-source impedance controller capable of minimising tracking errors in both task- and joint-space meaning that it is capable of performing precise pin-insertion tasks whilst remaining compliant and stable, see Fig.~\ref{fig:first-figure}. Furthermore, we are able to linearly interpolate between task- and joint-space control seamlessly. We demonstrate the controller capabilities on our Clearpath Ridgeback with two Kinova Gen3 manipulators christened Frank. We foresee that this controller will accurately track trajectories generated using both learning- and planning-based methods.

%% file: sections/2_methods.tex
\section{Impedance Formulation}

\begin{figure*}[t]
    \centering
    \includegraphics[width=\linewidth]{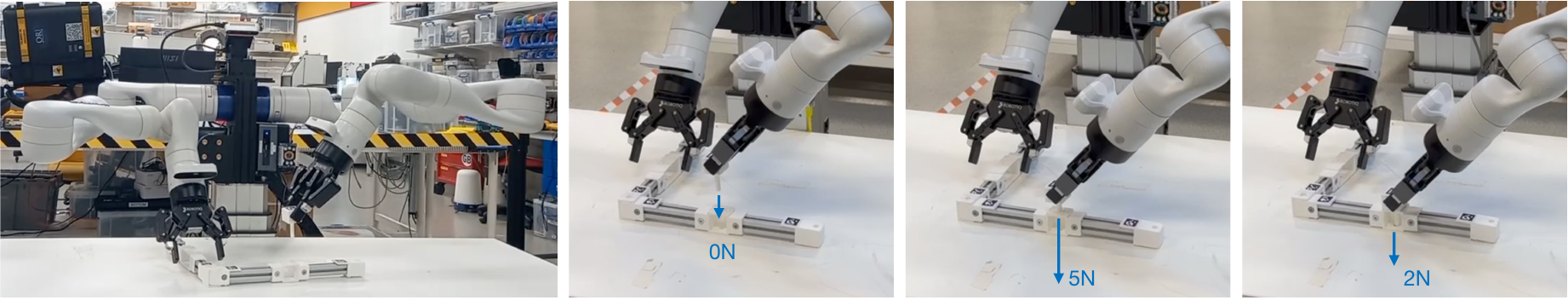}
    \caption{The dual-arm robot performing a precise pin insertion task during a RAMP~\cite{collins2023ramp} small-batch manufacturing task.
    The end-effector forces are measured by projecting the joint torques into the end-effector frame. Complete videos are found \href{https://github.com/applied-ai-lab/compliant_controllers}{here}.}
    \label{fig:results_figure}
    \vspace{-0.5cm}
\end{figure*}

Impedance control models the robot’s joints as torsional spring-dampers making them compliant. This compliance improves the safety of our robot when interacting with people and its environment. The end-effector tracking error is reduced using a virtual linear spring-damper system at the end-effector frame. To further reduce task-space errors, dynamics and static friction are reduced using a model-free observer. These objectives form the controller and their formulation are introduced in \cite{compliant_formulation}. In the following paragraphs, we introduce each component of our controller in turn.

For task- and joint-space compliant control, the motor-side joint torque $\tau_m$ is the sum of the joint-space positional and differential (PD) torque objective $\tau_q$, the task-space torque $\tau_x$ less the observed torque resulting from friction estimated by the friction observer $\hat{\tau}_f$ summed with torques to account for gravity compensation $g(q^*)$. This creates the following objective:
\begin{align}\label{eq:torque_m}
    \tau_m = \tau_q + \tau_x - \hat{\tau}_f + g(q^*)
\end{align}

The joint-space torque is evaluated using the well-known PD control law
\begin{align}\label{eq:joint_space_control}
    \tau_q = -K_{p} (q - q^*) - K_d (\dot{q} - \dot{q}^*),
\end{align}
where $K_p$ and $K_d$ are the position and derivative gains, $q$ is the current joint angles, $q^*$ are the desired or target joint angles, $\dot{q}$ are joint velocities. 

The task-space tracking PD controller is similar to the joint-space controller:
\begin{align}
    \tau_x = J^T(q) ( -K_{px} (x - x^*) - K_{dx} (\dot{x} - \dot{x}^*)).
\end{align}
The end-effector and desired poses are $x$ and $x^*$, whilst velocity terms are $\dot{x}$. The matrices $K_{px}$ and $K_{dx}$ are the position and derivative gains defined in task space. The Jacobian transposed $J^T$ projects the task-space wrench into joint-space torques applied at the motor side.

To guarantee stability and asymptotic convergence, both joint- and task-space controllers can be made passive by setting $\dot{q}^*$ and $\dot{x}^*$ to zero~\cite{Tiseo2023Safe}. 
However, this may limit velocity tracking performance therefore, bounding the maximum torque contribution from the derivative terms is preferred. The relative magnitudes of the $K_p$ and $K_{px}$ gains determine whether the controller joint- or task-space dominant.

The friction observer acts in joint-space and counteracts friction between the actuator's motor and link sides. The nominal joint angle $q_n$ is an estimate of the motor-side joint angles in absence of friction. This is estimated using the difference between the measured torque $\tau$ at the link side and the desired torque. The rotor inertia matrix $K_r$ is used to convert the torque difference into a joint acceleration
\begin{align}
    \ddot{q}_n = K_r^{-1} (\tau_m - \tau),
\end{align}
which is integrated twice to give the nominal joint angle $q_n$.

The friction observer acts to minimise the difference between the measured joint angles $q$ and the nominal $q_n$. The observer is formulated as
\begin{align}
    \hat{\tau}_f = K_r K_l \bigg((&\dot{q}_n - \dot{q}) \\
    &+ K_{lp} (q_n - q) \\ 
    &+ K_{li} \int_{t-T}^{t} (q_n - q) dt \bigg)
\end{align}
and is either a PD or PID controller depending on the values of the observer gains. The gain matrices $K_l$, $K_{lp}$, and $K_{li}$ are the observer gain, the proportional observer gain, and the integral observer gain. Setting the latter matrix to zero will convert the observer form a PID to a PD observer. 

Once the observer is used to estimate the nominal joint-angle, the PD control law in Eq.~\ref{eq:joint_space_control} is updated to 
\begin{align}
    \tau_q = -K_{p} (q_n - q^*) - K_d (\dot{q_n} - \dot{q}^*)
\end{align}
and the error between the nominal and desired joint position is minimised.

\section*{Implementation and Usage}

The controller is implemented in C++ using real-time programming techniques to ensure deterministic performance. All data containers are preallocated at construction time, and no heap allocations occur during runtime. The implementation is deployed on a PREEMPT\_RT real-time Linux kernel inside the ROS control framework \texttt{ros\_control}~\cite{chitta2017} as an extension of the popular \texttt{JointTrajectoryController} with an \texttt{EffortJointInterface}. 

The algorithmic code is separated from the ROS-Control implementation. This enables integration with ROS-based systems whilst providing the flexibility to use the algorithmic code for non-ROS users. Users can interface the controller with motion planners such as MoveIt!~\cite{coleman2014} or custom learning-based controllers. Unlike most impedance controllers, which require restarting or recompilation to switch control modes, our implementation provides a flexible runtime reconfiguration interface for seamless transitions between joint-only, task-only, or hybrid modes.

\begin{figure*}[t]
    \centering
    \includegraphics[width=1.0\linewidth]{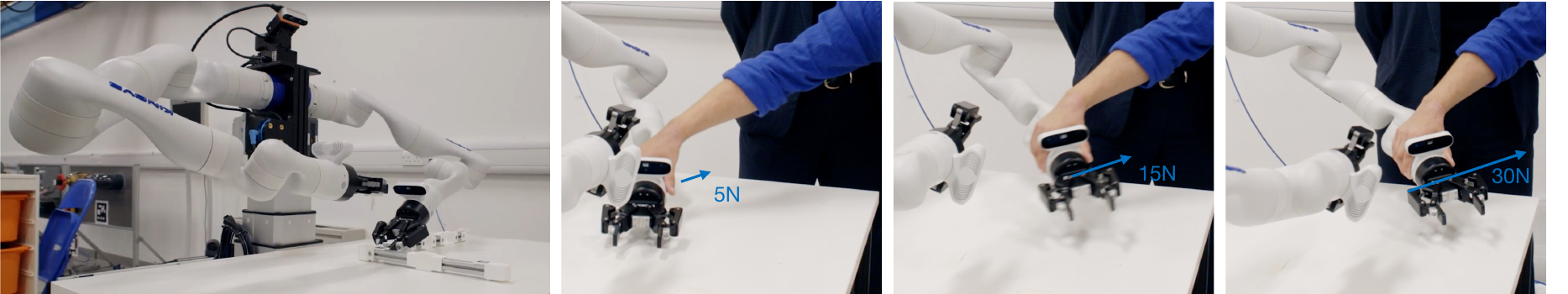}
    \caption{The controller is compliant, meaning that it is safe for humans to intervene during operation. Here, a bystander applies a force of up to 30N to the left robot arm. Simultaneously, the right arm reacts to this disturbance and tracks the disturbance.}
    \label{fig:intervention}
    \vspace{-0.5cm}
\end{figure*}

\subsection{Sending Setpoints To The Controller}

The \texttt{JointTrajectoryController} has two methods for sending trajectories. The first method is to publish a set point of type JointTrajectoryPoint, whilst the second is to publish a trajectory of type JointTrajectory to an action server. The first method is designed for streaming joint configurations to the controller for closed-loop control using either a planning- or learning-based method or for tele-operation. The latter interface also works for closed-loop control, but requires a trajectory in form of a list of JointTrajectoryPoint set points.

\subsection{Dynamic Reconfigure for Different Controller Behaviours}

A key feature of our system is the ability to adjust the contribution of joint- and task-space control at runtime. The controller ships with a \texttt{dynamic\_reconfigure} interface, allowing developers to modify gain profiles on the fly using \texttt{rqt\_reconfigure}, or programmatically via ROS publishers. The default configuration offers a balanced joint-task impedance profile tuned for safe interaction and accurate trajectory tracking.

Transitioning between modes is intuitive: reducing joint gains to zero creates a task-space-only controller, while zeroing task-space gains yields a joint-space-only controller. Intermediate profiles are realised using non-zero task and joint gains. This can be done continuously to blend between behaviours smoothly. This capability is particularly valuable for tasks that demand variable compliance, such as transitioning from broad-reaching motions to fine assembly insertions.




%% file: sections/3_Results.tex
\section{Results}

The task- and joint-space controller is deployed on our dual-arm manipulation platform Frank.
Frank is comprised of two Kinova Gen3 manipulators mounted along the horizontal axis, see Fig.~\ref{fig:results_figure}.
We emphasise that the evaluation of this controller is ongoing.
Nevertheless, we are satisfied with the reliability and performance of the controller on our dual-arm platform.

The controller set up in the experiments seen in Fig.~\ref{fig:first-figure} and~\ref{fig:results_figure} is operating at the joint and task level and is our default setting.
This mode results in a controller which is compliant enough to permit human interaction using finger-tip forces only as seen in Fig.\ref{fig:first-figure}, whilst remaining accurate enough to perform pin insertions where the pin diameter and hole are on the order of \SI{8}{\milli \meter}, see Fig.~\ref{fig:results_figure}. We emphasise that the gain profiles are identical during pin insertion and human interaction.

\subsection{Controller Evaluation}\label{sec:evaluation}

The controller is evaluated using the RAMP assembly benchmark~\cite{collins2023ramp}. The robot picks up a pin and inserts it into a hole. The locations of the pin and the hole are estimated using april tags on the RAMP installed constituent parts. The robot picks up the pin from its initial condition and places it in the hole to evaluate the controller's precision. This process is repeated ten times. Fig.~\ref{fig:results_figure} depicts the pin insertion and is annotated with representative force values measured by projecting the joint torque values into the end-effector's frame. Table~\ref{tab:rmse} shows the root mean squared error (RMSE) during pin insertion captured over a \SI{24}{\second} time window. These RMSE values are sufficiently small for successful pin insertion for all ten repetitions.

\begin{table}[h]
    \begin{center}
        \begin{tabular}{ | m{1.5cm} | m{1.5cm} | m{1.5cm} | m{1.5cm} | } 
          \hline
          & \textbf{X axis} & \textbf{Y axis} & \textbf{Z axis} \\
          \hline
          RMSE [cm]         &  0.41           & 0.53           & 1.20 \\
          \hline
        \end{tabular}
        \caption{The root mean squared error (RMSE) during pin insertion as seen in Fig.~\ref{fig:results_figure}. These values are calculated over a \SI{24}{\second} time window.}
        \label{tab:rmse}
    \end{center}
    \vspace{-0.5cm}
\end{table}

\subsection{Qualitative Results}

The controller is designed with compliance in mind for both assembly tasks and safe human interaction. To illustrate the controller's behaviour in response to human interaction, Fig.~\ref{fig:intervention} depicts the manipulator's response to an operator adversarially interacting with the robot's end-effector. In this figure, the operator applies a force of \SI{30}{\newton} to the arm. In Fig.~\ref{fig:first-figure}, we show the response of the task- and joint-space controller when an operator interacts with the robot's fourth joint, which is mid-way along the arm. Due to the redundancy of the seven degree-of-freedom arm, the end-effector pose remains unaltered, and once the fourth joint is released, the arm returns to the initial joint configuration.

%% file: sections/4_conclusions.tex
\section{Conclusions}

Impedance control renders rigid robots such as the Kinova Gen3 compliant. Impedance control at the joint level only introduces significant errors at the end-effector. To overcome this, we minimise the end-effector error by combining a task-space formulation with the joint-space one. Further end-effector tracking errors are minimised using a model-free friction observer introduced in~\cite{compliant_formulation}. This estimates the torque resulting from the static and dynamic friction. The resulting task- and joint-space controller is precise enough to perform pin-insertion tasks whilst being compliant enough for human interaction using finger-tip forces. The controller can be linearly switch between task and joint modes as well as operate in an intermediary state, which is the default setting. 
This controller is suitable for tracking motions from both learning-based and optimal-control methods.

%% file: sections/5_acknowledgements.tex
\section*{Acknowledgements}

This work was supported by a UKRI/EPSRC Programme Grant [EP/V000748/1].